\documentclass{article}

\usepackage[final, nonatbib]{neurips_2019}

\def\codename{CondConv}

\usepackage[utf8]{inputenc} %
\usepackage[T1]{fontenc}    %
\usepackage{hyperref}       %
\usepackage{url}            %
\usepackage{booktabs}       %
\usepackage{amsfonts}       %
\usepackage{nicefrac}       %
\usepackage{microtype}      %
\usepackage{graphicx}
\usepackage{subcaption}
\usepackage{amsmath}
\usepackage{appendix}

\title{CondConv: Conditionally Parameterized Convolutions for Efficient Inference}

\author{%
 Brandon Yang\thanks{Work done as part of the Google AI Residency.}\\
  Google Brain\\
  \texttt{bcyang@google.com} \\
   \And
   Gabriel Bender \\
   Google Brain \\
   \texttt{gbender@google.com} \\
   \AND
   Quoc V. Le \\
   Google Brain \\
   \texttt{qvl@google.com} \\
   \And
   Jiquan Ngiam \\
   Google Brain \\
   \texttt{jngiam@google.com} \\
}

\begin{document}

\maketitle

\begin{abstract}
Convolutional layers are one of the basic building blocks of modern deep neural networks. One fundamental assumption is that convolutional kernels should be shared for all examples in a dataset. We propose conditionally parameterized convolutions (\codename{}), which learn specialized convolutional kernels for each example. Replacing normal convolutions with \codename{} enables us to increase the size and capacity of a network, while maintaining efficient inference. We demonstrate that scaling networks with \codename{} improves the performance and inference cost trade-off of several existing convolutional neural network architectures on both classification and detection tasks. On ImageNet classification, our \codename{} approach applied to EfficientNet-B0 achieves state-of-the-art performance of 78.3\% accuracy with only 413M multiply-adds. Code and checkpoints for the \codename{} Tensorflow layer and \codename{}-EfficientNet models are available at: \url{https://github.com/tensorflow/tpu/tree/master/models/official/efficientnet/condconv}.

\end{abstract}

\section{Introduction}

Deep convolutional neural networks (CNNs) have achieved state-of-the-art performance on many tasks in computer vision~\cite{lecun1990handwritten,alexnet12}. Improvements in performance have largely come from increasing model size and capacity to scale to larger and larger datasets~\cite{mahajan2018exploring,gpipe18,amoebanets18}. However, current approaches to increasing model capacity are computationally expensive. Deploying the best-performing models for inference can consume significant datacenter capacity~\cite{jouppi2017datacenter} and are often not feasible for applications with strict latency constraints.

One fundamental assumption in the design of convolutional layers is that the same convolutional kernels are applied to every example in a dataset. To increase the capacity of a model, model developers usually add more convolutional layers or increase the size of existing convolutions (kernel height/width, number of input/output channels). In either case, the computational cost of additional capacity increases proportionally to the size of the input to the convolution, which can be large.

Due to this assumption and focus on mobile deployment, current computationally efficient models have very few parameters \cite{howard2017mobilenets, sandler2018mobilenetv2, tan2018mnasnet}. However, there is a growing class of computer vision applications that are not constrained by parameter count, but have strict latency requirements at inference, such as real-time server-side video processing and perception for self-driving cars. In this paper, we aim to design models to better serve these applications.

We propose conditionally parameterized convolutions (\codename{}), which challenge the paradigm of static convolutional kernels by computing convolutional kernels as a function of the input. In particular, we parameterize the convolutional kernels in a \codename{} layer as a linear combination of $n$ experts $(\alpha_1 W_1 + \ldots + \alpha_n W_n) * x$, where $\alpha_1, \ldots, \alpha_n$ are functions of the input learned through gradient descent. To efficiently increase the capacity of a \codename{} layer, model developers can increase the number of experts. This is much more computationally efficient than increasing the size of the convolutional kernel itself, because the convolutional kernel is applied at many different positions within the input, while the experts are combined only once per input. This allows model developers to increase model capacity and performance while maintaining efficient inference.

\codename{} can be used as a drop-in replacement for existing convolutional layers in CNN architectures. We demonstrate that replacing convolutional layers with \codename{} improves model capacity and performance on several CNN architectures on ImageNet classification and COCO object detection, while maintaining efficient inference. In our analysis, we find that \codename{} layers learn semantically meaningful relationships across examples to compute the conditional convolutional kernels.

\begin{figure}[t]
  \centering
  \begin{subfigure}[b]{0.50\textwidth}
    \centering\includegraphics[width=0.92\textwidth]{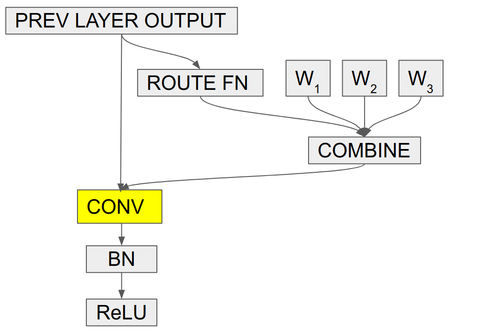}
    \caption{\label{fig:condconv_layer} \codename{}: $(\alpha_1 W_1 + \ldots + \alpha_n W_n) * x$}
  \end{subfigure}%
  \begin{subfigure}[b]{0.50\textwidth}
    \centering\includegraphics[width=0.96\textwidth]{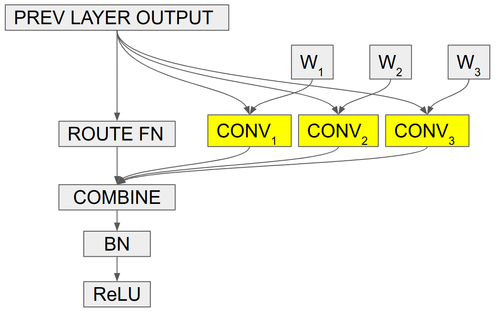}
    \caption{\label{fig:moe_layer} Mixture of Experts: $\alpha_1 (W_1 * x) + \ldots + \alpha_n (W_n * x)$}
  \end{subfigure}
  \caption{(\subref{fig:condconv_layer}) Our \codename{} layer architecture with $n=3$ kernels vs. (\subref{fig:moe_layer}) a mixture of experts approach. By parameterizing the convolutional kernel conditionally on the input, \codename{} is mathematically equivalent to the mixture of experts approach, but requires only $1$ convolution.}
  \label{fig:condconv_figure}
\end{figure}

\section{Related Work}

\textbf{Conditional computation}. 
Similar to \codename{}, conditional computation aims to increase model capacity without a proportional increase in computation cost. In conditional computation models, this is achieved by activating only a portion of the entire network for each example ~\cite{bengio2013estimating, davis2013low, cho2014exponentially, bengio2015conditional}. However, conditional computation models are often challenging to train, since they require learning discrete routing decisions from individual examples to different experts. Unlike these approaches, \codename{} does not require discrete routing of examples, so can be easily optimized with gradient descent.

One approach to conditional computation uses reinforcement learning or evolutionary methods to learn discrete routing functions~\cite{rosenbaum2017routing, mcgill2017deciding, liu2018dynamic, fernando2017pathnet}. BlockDrop~\cite{wu2018blockdrop} and SkipNet~\cite{wang2018skipnet} use reinforcement learning to learn the subset of blocks needed to process a given input. Another approach uses unsupervised clustering methods to partition examples into sub-networks. Gross et al.~\cite{gross2017hard} use a two stage training and clustering pipeline to train a hard mixture of experts model. Mullapudi et al.~\cite{ravi18} use clusters as labels to train a routing function between branches in a deep CNN model. Finally, Shazeer et al.~\cite{shazeer2017outrageously} proposed the sparsely-gated mixture-of-experts layer, which which achieves significant success on large language modeling using noisy top-k gating.
 
Prior work in computation demonstrates the potential of designing large models that process different sets of examples with different sub-networks. Our work on \codename{} pushes the boundaries of this paradigm, by enabling each individual example to be processed with different weights.

\textbf{Weight generating networks}. Ha et al. \cite{ha2016hypernetworks} propose the use of a small network to generate weights for a larger network. Unlike \codename{}, for CNNs, these weights are the same for every example in the dataset. This enables greater weight-sharing, which achieves lower parameter count but worse performance than the original network. In neural machine translation, Platanios et al. \cite{platanios2018contextual} generate weights to translate between different language pairs, but use the same weights for every example within each language pair. 

\textbf{Multi-branch convolutional networks}. Multi-branch architectures like Inception~\cite{szegedy2015going} and ResNext~\cite{xie2017aggregated} have shown success on a variety of computer vision tasks. In these architectures, a layer consists of multiple convolutional branches, which are aggregated to compute the final output. A \codename{} layer is mathematically equivalent to a multi-branch convolutional layer where each branch is a single convolution and outputs are aggregated by a weighted sum, but only requires the computation of one convolution.

\textbf{Example dependent activation scaling}. Some recent work proposes to adapt the activations of neural networks conditionally on the input. Squeeze-and-Excitation networks~\cite{hu2018squeeze} learn to scale the activations of every layer output. GaterNet~\cite{chen18} uses a separate network to select a binary mask for filters for a larger backbone network. Attention-based methods ~\cite{luong2015effective, bahdanau2014neural, vaswani2017attention} scale previous layer inputs based on learned attention weights. Scaling activations has similar motivations as \codename{}, but is restricted to modulating activations in the base network.

\textbf{Input-dependent convolutional layers}. In language modeling, Wu et al. \cite{wu2019pay} use input-dependent convolutional kernels as a form of local attention. In vision, Brabandere et al.~\cite{jia2016dynamic} generate small input-dependent convolutional filters to transform images for next frame and stereo prediction. Rather than learning input-dependent weights, Dai et al.~\cite{dai2017deformable} propose to learn different convolutional offsets for each example. Finally, in recent work,  SplineNets~\cite{keskin18} apply input-dependent convolutional weights, modeled as 1-dimensional B-splines, to implement continuous neural decision graphs.

\section{Conditionally Parameterized Convolutions}
In a regular convolutional layer, the same convolutional kernel is used for all input examples. In a \codename{} layer, the convolutional kernel is computed as a function of the input example (Fig \ref{fig:condconv_layer}). Specifically, we parameterize the convolutional kernels in \codename{} by:
$$ Output(x) = \sigma((\alpha_1 \cdot W_1 + \ldots + \alpha_n \cdot W_n) * x) $$
where each $\alpha_i = r_i(x)$ is an example-dependent scalar weight computed using a routing function with learned parameters, $n$ is the number of experts, and $\sigma$ is an activation function. When we adapt a convolutional layer to use \codename{}, each kernel $W_i$ has the same dimensions as the kernel in the original convolution.

We typically increase the capacity of a regular convolutional layer by increasing the kernel height/width or number of input/output channels. However, each additional parameter in a convolution requires additional multiply-adds proportional to the number of pixels in the input feature map, which can be large. In a \codename{} layer, we compute a convolutional kernel for each example as a linear combination of $n$ experts before applying the convolution. Crucially, each convolutional kernel only needs to be computed once but is applied at many different positions in the input image. This means that by increasing $n$, we can increase the capacity of the network with only a small increase in inference cost; each additional parameter requires only 1 additional multiply-add.

A \codename{} layer is mathematically equivalent to a more expensive linear mixture of experts formulation, where each expert corresponds to a static convolution (Fig \ref{fig:moe_layer}):
$$ \sigma((\alpha_1 \cdot W_1 + \ldots + \alpha_n \cdot W_n) * x) = \sigma(\alpha_1 \cdot (W_1 * x) + \ldots + \alpha_n \cdot (W_n * x)) $$
Thus, \codename{} has the same capacity as a linear mixture of experts formulation with $n$ experts, but is computationally efficient since it requires computing only one expensive convolution. This formulation gives insight into the properties of \codename{} and relates it to prior work on conditional computation and mixture of experts. The per-example routing function is crucial to \codename{} performance: if the learned routing function is constant for all examples, a \codename{} layer has the same capacity as a static convolutional layer.

We wish to design a per-example routing function that is computationally efficient, able to meaningfully differentiate between input examples, and is easily interpretable. We compute the example-dependent routing weights $\alpha_i = r_i(x)$ from the layer input in three steps: global average pooling, fully-connected layer, Sigmoid activation. 
$$ r(x) = \text{Sigmoid}(\text{GlobalAveragePool}(x) \: R) $$
where $R$ is a matrix of learned routing weights mapping the pooled inputs to $n$ expert weights. A normal convolution operation operates only over local receptive fields, so our routing function allows adaptation of local operations using global context.

The \codename{} layer can be used in place of any convolutional layer in a network. The same approach can easily be extended to other linear functions like those in depth-wise convolutions and fully-connected layers.

\section{Experiments} \label{experiments}

We evaluate \codename{} on ImageNet classification and COCO object detection by scaling up the MobileNetV1 ~\cite{howard2017mobilenets}, MobileNetV2 ~\cite{sandler2018mobilenetv2}, ResNet-50 ~\cite{he2016deep}, MnasNet ~\cite{tan2018mnasnet}, and EfficientNet ~\cite{tan2019efficientnet} architectures. In practice, we have two options to train \codename{} models, which are mathematically equivalent. We can either first compute the kernel for each example and apply convolutions with a batch size of one (Fig. \ref{fig:condconv_layer}), or we can use the linear mixture of experts formulation (Fig. \ref{fig:moe_layer}) to perform batch convolutions on each branch and sum the outputs. Current accelerators are optimized to train on large batch convolutions, and it is difficult to fully utilize them for small batch sizes. Thus, with small numbers of experts (<=4), we found it to be more efficient to train \codename{} layers with the linear mixture of experts formulation and large batch convolutions, then use our efficient \codename{} approach for inference. With larger numbers of experts (>4), training \codename{} layers directly with batch size one is more efficient.

\subsection{ImageNet Classification}

\begin{figure}[t]
\centering
\includegraphics[width=0.6\textwidth]{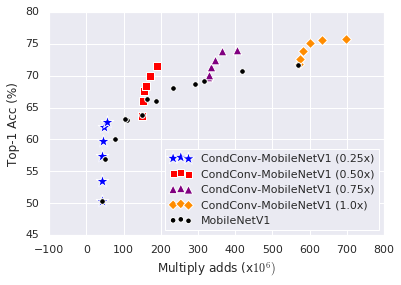}
\caption{On ImageNet validation, increasing the number of experts per layer of our \codename{}-MobileNetV1 models improves performance relative to inference cost compared to the MobileNetV1 frontier~\cite{slim2016} across a spectrum of model sizes. Models with more experts per layer achieve monotonically higher accuracy. We train \codename{} models with \{1, 2, 4, 8, 16, 32\} experts at width multipliers \{0.25, 0.50, 0.75, 1.0\}.}
\label{fig:ccmv1v1_acc_madds}
\end{figure}

We evaluate our approach on the ImageNet 2012 classification dataset~\cite{russakovsky2015imagenet}. The ImageNet dataset consists of 1.28 million training images and 50K validation images from 1000 classes. We train all models on the entire training set and compare the single-crop top-1 validation set accuracy with input image resolution 224x224. For MobileNetV1, MobileNetV2, and ResNet-50, we use the same training hyperparameters for all models on ImageNet, following \cite{kornblith2018better}, except we use BatchNorm momentum of 0.9 and disable exponential moving average on weights. For MnasNet~\cite{tan2018mnasnet} and EfficientNet~\cite{tan2019efficientnet}, we use the same training hyperparameters as the original papers, with the batch size, learning rate, and training steps scaled appropriately for our hardware configuration. For fair comparison, we retrain all of our baseline models with the same hyperparameters and regularization search space as the \codename{} models. We used accuracy on the validation set to determine early stopping. We measure performance as ImageNet top-1 accuracy relative to computational cost in multiply-adds (MADDs).

For each baseline architecture, we evaluate \codename{} by replacing convolutional layers with \codename{} layers, and increasing the number of experts per layer. We share routing weights between layers in a block (a residual block, inverted bottleneck block, or separable convolution). Additionally, for some models, we replace the fully-connected classification layer with a 1x1 \codename{} layer. For the exact architectural details, refer to Appendix A. Our ablation experiments in Table \ref{table:ablation_gating_fns} and Table \ref{table:ablation_mixindexbegin} suggest \codename{} improves performance across a wide range of layer and routing architectures.

We use two general regularization techniques for models with large capacity. First, we use Dropout~\cite{srivastava2014dropout} on the input to the fully-connected layer preceding the logits, with keep probability between 0.6 and 1.0. Second, we also add data augmentation using the AutoAugment~\cite{cubuk2018autoaugment} ImageNet policy and Mixup~\cite{zhang2017mixup} with $\alpha=0.2$. To address overfitting in the large ResNet models, we additionally introduce a new data augmentation technique for \codename{} based on Shake-Shake~\cite{gastaldi2017shake} by randomly dropping out experts during training. 

\begin{table}[t]
    \centering
    \caption{ImageNet validation accuracy and inference cost for our \codename{} models on several baseline model architectures. All models use 8 experts per \codename{} layer. \codename{} improves the accuracy of all baseline architectures with small relative increase in inference cost (<10\%).}
    \begin{center}
        \begin{tabular}{c  @{\hspace{3.5\tabcolsep}} c  @{\hspace{1.5\tabcolsep}} c  @{\hspace{3.5\tabcolsep}} c  @{\hspace{1.5\tabcolsep}} c} 
        \toprule
         & \multicolumn{2}{c}{Baseline} & \multicolumn{2}{c}{\codename{}} \\
         & MADDs ($\times 10^6$) & Top-1 (\%) &  MADDs ($\times 10^6$) &  Top-1 (\%)\\
        \midrule
        MobileNetV1 (1.0x) & 567 & 71.9 & 600 & 73.7 \\
        MobileNetV2 (1.0x) & 301 & 71.6 & 329 & 74.6 \\
        MnasNet-A1 & 312 & 74.9 & 325 & 76.2 \\
        ResNet-50 & 4093 & 77.7 & 4213 & 78.6 \\
        EfficientNet-B0 & 391 & 77.2 & 413 & 78.3 \\
        \bottomrule
        \end{tabular}
    \end{center}
    \label{table:ccmv2_imagenet}
\end{table}

\stepcounter{footnote}\footnotetext{Our re-implementation of the baseline models and our \codename{} models use the same hyperparameters and regularization search space for fair comparison. For reference, published results for baselines are:\\MobileNetV1 (1.0x): 70.6\%~\cite{howard2017mobilenets}. MobileNetV2 (1.0x): 72.0\%~\cite{sandler2018mobilenetv2}. MnasNet-A1: 75.2\%~\cite{tan2018mnasnet}. ResNet-50: 76.4\%~\cite{goyal2017accurate}. EfficientNet-B0: 76.3\%~\cite{tan2019efficientnet}.}

On MobileNetV1, we find that increasing the number of \codename{} experts improves accuracy relative to inference cost compared to the performance frontier with static convolutional scaling techniques using the channel width and input size (Figure \ref{fig:ccmv1v1_acc_madds}). Moreover, we find that increasing the number of \codename{} experts leads to monotonically increasing performance with sufficient regularization.

We further find that \codename{} improves performance relative to inference cost on a wide range of architectures (Table \ref{table:ccmv2_imagenet}). This includes architectures that take advantage of architecture search~\cite{tan2019efficientnet, tan2018mnasnet}, Squeeze-and-Excitation~\cite{hu2018squeeze}, and large architectures with ordinary convolutions not optimized for inference time~\cite{he2016deep}. For more in depth comparisons, see Appendix A. 

Our \codename{}-EfficientNet-B0 model achieves state-of-the-art performance of 78.3\% accuracy with 413M multiply-adds, when compared to the MixNet frontier~\cite{tan2019mixnet}. To directly compare our \codename{} scaling approach to the compound scaling coefficient proposed by Tan et al.~\cite{tan2019efficientnet}, we additionally scale the \codename{}-EfficientNet-B0 model with a depth multiplier of 1.1x, which we call \codename{}-EfficientNet-B0-depth. Our \codename{}-EfficientNet-B0-depth model achieves 79.5\% accuracy with only 614M multiply-adds. When trained with the same hyperparameters and regularization search space, the EfficientNet-B1 model, which is scaled from the EfficientNet-B0 model using the compound coefficient, achieves 79.2\% accuracy with 700M multiply-adds. In this regime, \codename{} scaling outperforms static convolutional scaling with the compound coefficient.

\subsection{COCO Object Detection}

\begin{table}[h]
    \centering
    \caption{COCO object detection minival performance of our \codename{}-MobileNetV1 SSD 300 architecture with 8 experts per layer. Mean average precision (mAP) reported with COCO primary challenge metric (AP at IoU=0.50:0.05:0.95). \codename{} improves mAP at all model sizes with small relative increase in inference cost (<5\%).}
    \begin{center}
        \begin{tabular}{c  @{\hspace{3.5\tabcolsep}} c  @{\hspace{1.5\tabcolsep}} c  @{\hspace{3.5\tabcolsep}} c  @{\hspace{1.5\tabcolsep}} c} 
        \toprule
         & \multicolumn{2}{c}{Baseline} & \multicolumn{2}{c}{\codename{}} \\
         & MADDs ($\times 10^6$) & mAP & MADDs ($\times 10^6$) & mAP \\
         \midrule
        MobileNetV1 (0.5x) & 352 & 14.4 & 363 & 18.0 \\
        MobileNetV1 (0.75x) & 730 & 18.2 & 755 & 21.0 \\
        MobileNetV1(1.0x) & 1230 & 20.3 & 1280 & 22.4 \\
        \bottomrule
        \end{tabular}
    \end{center}
    \label{table:detection}
\end{table}

\begin{table}[t]
   \begin{minipage}{0.48\textwidth}
     \centering
        \caption{Different routing architectures. Our baseline \codename{}(CC)-MobileNetV1 uses a one-layer, fully-connected routing function with Sigmoid activation for each \codename{} block.\\
        \label{table:ablation_gating_fns}}
        \resizebox{1.0\textwidth}{!}{\begin{tabular}{c c c c c c} 
        \toprule
        Routing Fn & MADDs & Valid \\
         & ($\times 10^6$) &  Top-1 (\%) \\
        \midrule
        CC-MobileNetV1 (0.25x) & 55.7 & 62.0 \\
        \midrule
        Single & 55.5 & 56.5 \\
        Partially-shared & 55.6 & 62.5 \\
        \midrule
        Hidden (small) & 55.6 & 57.7 \\
        Hidden (medium) & 55.9 & 62.2 \\
        Hidden (large) & 57.8 & 54.1 \\
        \midrule
        Hierarchical & 55.7 & 60.3 \\
        \midrule
        Softmax & 55.7 & 60.5 \\
        \bottomrule
        \end{tabular}}
   \end{minipage}\hfill
   \begin{minipage}{0.48\textwidth}
    \centering
    \begin{center}
        \caption{\codename{} at different layers in our \codename{}(CC)-MobileNetV1 (0.25x) model. FC refers to the final classification layer. \codename{} improves performance at every layer.\\   \label{table:ablation_mixindexbegin}}
        \resizebox{1.0\textwidth}{!}{\begin{tabular}{c c c c} 
        \toprule
        \codename{} Begin & MADDs & Valid \\
         Layer &  ($\times 10^6$) & Top-1 (\%) \\
        \midrule
        CC-MobileNetV1 (0.25x) & 55.7 & 62.0 \\
        MobileNetV1 (0.25x) & 41.2 & 50.0 \\
        \midrule
        1 & 56.3 & 62.5 \\
        5 & 56.0 & 62.0 \\
        7 & 55.7 & 62.0 \\
        13 & 52.5 & 59.5 \\
        15 (FC Only) & 49.3 & 54.2 \\
        \midrule
        7 (No FC) & 47.6 & 60.2 \\
        \bottomrule
        \end{tabular}}
    \end{center}
   \end{minipage}
\end{table}

We next evaluate the effectiveness of \codename{} on a different task and dataset with the COCO object detection dataset~\cite{lin2014microsoft}. Our experiments use the MobileNetV1 feature extractor and the Single Shot Detector~\cite{liu2016ssd} with 300x300 input resolution (SSD300).

Following Howard et al.~\cite{howard2017mobilenets}, we train on the combined COCO training and validation sets excluding 8,000 minival images, which we evaluate our networks on. We train our models using a batch size of 1024 for 20,000 steps. For the learning rate, we use linear warmup from 0.3 to 0.9 over 1,000 steps, followed by cosine decay~\cite{loshchilov2016sgdr} from 0.9. We use the data augmentation scheme proposed by Liu et al.~\cite{liu2016ssd}. We use the same convolutional feature layer dimensions, SSD hyperparameters, and training hyperparameters across all models. We measure performance as COCO minival mean average precision (mAP) relative to computational cost in multiply-adds (MADDs).

We use our \codename{}-MobileNetV1 models with depth multipliers \{0.50, 0.75, 1.0\} as the feature extractors for object detection. We further replace the additional convolutional feature extractor layers in SSD with \codename{} layers.

\codename{} with 8 experts improves object detection performance at all model sizes (Table~\ref{table:detection}). Our \codename{}-MobileNetV1(0.75x) SSD model exceeds the MobileNetV1(1.0x) SSD baseline by 0.7 mAP at 60\% of the inference cost. Moreover, our \codename{}-MobileNetV1{}(1.0x) SSD model improves upon the MobileNetV1(1.0x) SSD baseline by 2.1 mAP at similar inference cost.

\stepcounter{footnote}\footnotetext{Our re-implementation of the baseline models and our \codename{} models use the same hyperparameters for fair comparison. As published reference, Howard et al.~\cite{howard2017mobilenets} report mAP of 19.3 for MobileNetV1 (1.0x).}

\subsection{Ablation studies} \label{ablation}

We perform ablation experiments to better understand model design with the \codename{} block. In all experiments, we compare against the same baseline \codename{}-MobileNetV1 (0.25x) model with 32 experts per \codename{} layer, trained with the same setup as Section \ref{experiments} and no additional Dropout or data augmentation. The baseline model achieves 61.98\% ImageNet Top-1 validation accuracy with 55.7M multiply-adds. The MobileNetV1(0.25x) architecture achieves 50.4\% Top-1 accuracy with 41.2M multiply-adds.\footnotemark[4] We choose this setup for ease of training and large effect of \codename{}.

\stepcounter{footnote}\footnotetext{Our implementation. Howard et al. \cite{howard2017mobilenets} report a top-1 accuracy of 50.0\% with different hyperparameters.}

\subsubsection{Routing function}

We investigate different choices for the routing function in Table \ref{table:ablation_gating_fns}. The baseline model computes new routing weights for each layer. \textit{Single} computes the routing weights only once at \codename{} 7 (the 7th separable convolutional block), and uses the same routing weights in all subsequent layers. \textit{Partially-shared} shares the routing weights between every other layer. Both the baseline model and \emph{Partially-shared} significantly outperform \emph{Single}, which suggests that routing at multiple depths in the network improve quality. \emph{Partially-shared} performs slightly outperforms the baseline, suggesting that sharing routing functions among nearby layers can improve quality.

We then experiment with more complex routing functions, by introducing a hidden layer with ReLU activation after the global average pooling step. We vary the hidden layer size to be $input\_dim / 8$ for \textit{Hidden (small)}, $input\_dim$ for \textit{Hidden (medium)}, and $input\_dim \cdot 8$ for \textit{Hidden (large)}. Adding a non-linear hidden layer of appropriate size can slightly improve performance. Large hidden layer sizes are prone to over-fitting, even with the same number of experts.

Next, we experiment with \textit{Hierarchical} routing functions, by concatenating the routing weights of the previous layer to the output of the global average pooling layer in the routing function. This adds a dependency between \codename{} routing weights, which we find is also prone to overfitting.

Finally, we experiment with the \textit{Softmax} activation function to compute routing weights. The baseline's \textit{Sigmoid} significantly outperforms \textit{Softmax}, which suggests that multiple experts are often useful for a single example.

\begin{figure*}[t]%
\centering
\begin{subfigure}[b]{0.3\textwidth}
\centering\includegraphics[width=\textwidth]{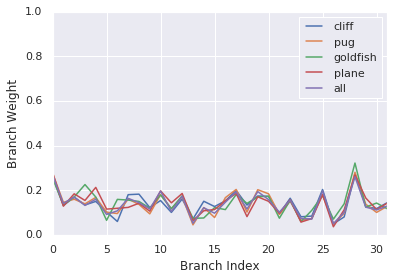}
\caption{\label{fig:layer7} Layer 12}
\end{subfigure}%
\begin{subfigure}[b]{0.3\textwidth}
\centering\includegraphics[width=\textwidth]{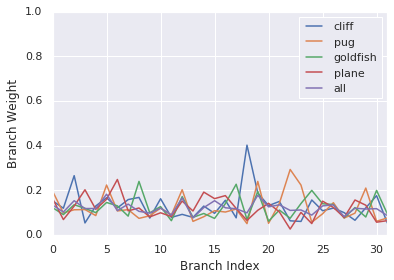}
\caption{\label{fig:layer13} Layer 26}
\end{subfigure}%
\begin{subfigure}[b]{0.3\textwidth}
\centering\includegraphics[width=\textwidth]{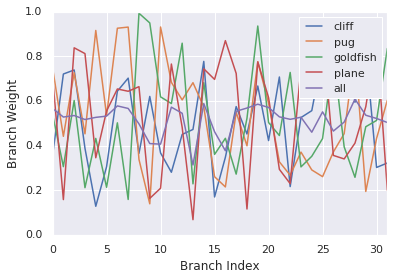}
\caption{\label{fig:layer14} Fully Connected (FC)}
\end{subfigure}%
\caption{Mean routing weights for four classes averaged across the ImageNet validation set at three different depths in our \codename{}-MobileNetV1 (0.5x) model. \codename{} routing weights are more class-specific at greater depth.}
\label{fig:branch_weights_layer}

\bigskip

\centering
\includegraphics[width=0.45\textwidth]{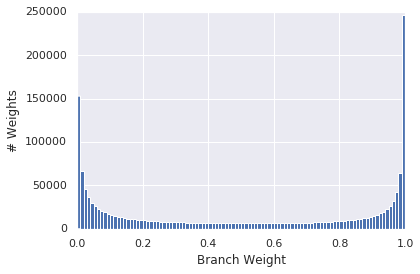}
\caption{Distribution of routing weights in the final \codename{} layer of our \codename{}-MobileNetV1 (0.5x) model when evaluated on all images in the ImageNet validation set. Routing weights follow a bi-modal distribution.}
\label{fig:branch_hist}
\end{figure*}

\subsubsection{\codename{} Layer Depth}

We analyze the effect of \codename{} layers at different depths in the \codename{}-MobileNetV1 (0.25x) model (Table \ref{table:ablation_mixindexbegin}). We use \codename{} layers in the begin layer, and all subsequent layers. We further perform ablation studies specific to the final fully-connected classification layer. We find \codename{} layers improve performance when applied at every layer in the network. Additionally, we find that additionally applying \codename{} before layer 7 in the network has only small effects on performance. For image classification with the \codename{}-MobileNetV1 (0.25x) model, \codename{} in the final classification layer accounts for a significant fraction of the additional inference cost. Using a normal final classification layer results in smaller performance gains, but is more efficient.

\section{Analysis} \label{analysis}

\begin{figure*}[t]%
\centering
\includegraphics[width=0.35\textwidth]{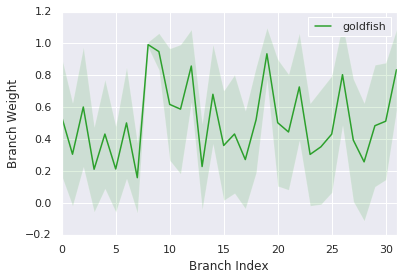}
\includegraphics[width=0.35\textwidth]{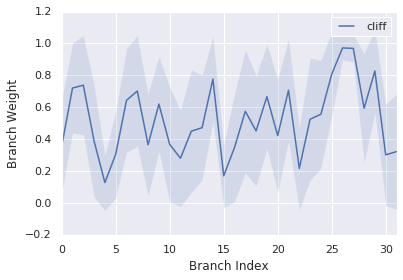}
\caption{Routing weights in the final \codename{} layer in our \codename{}-MobileNetV1 (0.5x) model for 2 classes averaged across the ImageNet validation set. Error bars indicate one standard deviation.}
\label{fig:perclass_branch_weights}

\bigskip

\centering
\includegraphics[width=\textwidth]{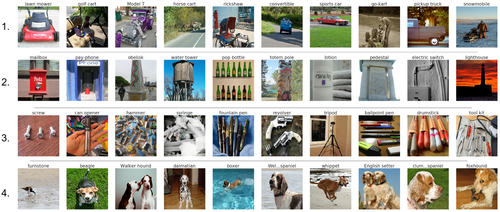}
\caption{Top 10 classes with highest mean routing weight for 4 different experts in the final \codename{} layer in our \codename{}-MobileNetV1 (0.5x) model, as measured across the ImageNet validation set. Expert 1 is most activated for wheeled vehicles; expert 2 is most activated for rectangular structures; expert 3 is most activated for cylindrical household objects; expert 4 is most activated for brown and black dog breeds.}
\label{fig:branch_top10}
\end{figure*}

In this section, we aim to gain a better understanding of the learned kernels and routing functions in our \codename{}-MobileNetV1 architecture. We study our \codename{}-MobileNetV1 (0.50x) architecture with 32 experts per layer trained on ImageNet with Mixup and AutoAugment, which achieves 71.6\% top-1 validation accuracy. We evaluate our \codename{}-MobileNetV1 (0.5x) model on the 50,000 ImageNet validation examples, and compute the routing-weights at \codename{} layers in the network.

We first study inter-class variation between the routing weights at different layers in the network. We visualize the average routing weight for four different classes (cliff, pug, goldfish, and plane, as suggested by Hu et al.~\cite{hu2018squeeze} for semantic and appearance diversity) at three different depths in the network (Layer 12, Layer 26, and the final fully-connected layer). The distribution of the routing weights is very similar across classes at early layers in the network, and become more and more class specific at later layers (Figure \ref{fig:branch_weights_layer}). This suggests an explanation for why replacing additional convolutional layers with \codename{} layers near the input of the network does not significantly improve performance.

We next analyze the distribution of the routing weights of the final fully-connected layer in Figure \ref{fig:branch_hist}. The routing weights follow a bi-modal distribution, with most experts receiving a routing weight close to 0 or 1. This suggests that the experts are sparsely activated, even without regularization, and further suggests the specialization of the experts.

We then study intra-class variation between the routing weights in the final \codename{} layer (Figure \ref{fig:perclass_branch_weights}). Within one class, some kernels are activated with high weight and small variance for all examples. However, even within one class, there can be big variation in the routing weights between examples.

Finally, to better understand experts in the final \codename{} layer, we visualize top 10 classes with highest mean routing weight for four difference experts on the ImageNet validation set (Figure \ref{fig:branch_top10}). We show the exemplar image with highest routing weight within each class. \codename{} layers learn to specialize in semantically and visually meaningful ways.

\section{Conclusion}

In this paper, we proposed conditionally parameterized convolutions (\codename{}). \codename{} challenges the assumption that convolutional kernels should be shared across all input examples. This introduces a new direction for increasing model capacity while maintaining efficient inference: increase the size and complexity of the kernel-generating function. Since the kernel is computed only once, then convolved across the input, increasing the complexity of the kernel-generating function can be much more efficient than adding additional convolutions or expanding existing convolutions. \codename{} also highlights an important research question in the trend towards larger datasets on how to best uncover, represent, and leverage the relationship between examples to improve model performance. In the future, we hope to further explore the design space and limitations of \codename{} with larger datasets, more complex kernel-generating functions, and architecture search to design better base architectures.

\bibliography{bibliography}
\bibliographystyle{plain}

\newpage

\appendix
\appendixpage
\section{ImageNet Architectures}

In this section, we provide detailed descriptions of the specific \codename{} architectures we used for each baseline model, and elaborate on individual results. All \codename{} results are reported with 8 experts per layer. For fair comparison, all results reported use the same training hyperparameters and regularization search space as the \codename{} model they are compared against.

\textbf{\codename{}-MobileNetV1}. We replace the convolutional layers starting from the sixth separable convolutional block and the final fully-connected classification layer of the baseline MobileNetV1 model with \codename{} layers. We share routing weights between depthwise and pointwise layers with a separable convolution block. Our \codename{}-MobileNetV1 (0.5x) model with 32 experts per \codename{} layer achieves 71.6\% accuracy at 190M multiply-adds, comparable to the MobileNetV1 (1.0x) model at 71.7\% at 571M multiply-adds.

\textbf{\codename{}-MobileNetV2}. We replace the convolutional layers in the final 6 inverted residual blocks and the final fully-connected classification layer of the baseline MobileNetV2 architecture with \codename{} layers. We share routing weights between convolutional layers in each inverted bottleneck block. Our \codename{}-MobileNetV2 (1.0x) model achieves 74.6\% accuracy at 329M multiply-adds. The MobileNetV2 (1.4x) architecture with static convolutions scaled by width multiplier achieves similar accuracy of 74.5\% in our implementation (74.7\% in \cite{sandler2018mobilenetv2}), but requires 585M multiply-adds.

\textbf{\codename{}-MnasNet-A1}. We replace the convolutional layers in the final 3 block groups of the baseline MnasNet-A1 architecture with \codename{} layers. We share routing weights between convolutional layers in each inverted bottleneck block within a block group. The baseline MnasNet-A1 model achieves 74.9\% accuracy with 312M multiply-adds. Our \codename{}-MnasNet-A1 model achieves 76.2\% accuracy with 329M multiply-adds. A larger model from the same search space using static convolutional layers, MnasNet-A2, achieves 75.6\% accuracy with 340M multiply-adds \cite{tan2018mnasnet}.

\textbf{\codename{}-ResNet-50}. We replace the convolutional layers in the final 3 residual blocks and the final fully-connected classification layer of the baseline ResNet-50 architecture with \codename{} layers. The baseline ResNet-50 model achieves 77.7\% accuracy at 4096M multiply-adds. Our \codename{}-ResNet-50 architecture achieves 78.6\% accuracy at 4213 multiply-adds. With sufficient regularization, \codename{} improves the performance of even large model architectures with ordinary convolutions that are not optimized for inference time.

\textbf{\codename{}-EfficientNet-B0}. We replace the convolutional layers in the final 3 block groups of the baseline EfficientNet-B0 architecture with \codename{} layers. We share routing weights between convolutional layers in each inverted bottleneck block within a block group. The baseline EfficientNet-B0 model achieves 77.2\% accuracy with 391M multiply-adds. Our \codename{}-EfficientNet-B0 model achieves 78.3\% accuracy with 413M multiply-adds. 

To directly compare our \codename{} scaling approach to the compound scaling coefficient proposed by Tan et al.~\cite{tan2019efficientnet}, we additionally scale the \codename{}-EfficientNet-B0 model with a depth multiplier of 1.1x, which we call \codename{}-EfficientNet-B0-depth. Our \codename{}-EfficientNet-B0-depth model achieves 79.5\% accuracy with only 614M multiply-adds. When trained with the same hyperparameters and regularization search space, the EfficientNet-B1 model, which is scaled from the EfficientNet-B0 model using the compound coefficient, achieves 79.2\% accuracy with 700M multiply-adds. In this regime, \codename{} scaling outperforms static convolutional scaling with the compound coefficient.

\end{document}